# FGSGT: Saliency-Guided Siamese Network Tracker Based on Key Fine-Grained Feature Information for Thermal Infrared Target Tracking

Ruoyan Xiong[1,2][†], Huanbin Zhang[1,2][†], Shentao Wang[3][*], Hui He[4], Yuke Hou[1,2], Yue Zhang[1,2], Yujie Cui[1,2], Huipan Guan[1,2], and Shang Zhang[1,2]

[1] College of Computer and Information Technology, China Three Gorges University, Yichang 443002, China
[2] Hubei Province Engineering Technology Research Center for Construction Quality Testing Equipment, China Three Gorges University, Yichang 443002, China
[3] Nantong Institute of Technology, Nantong 226001, China
[4] Wuhan College, Wuhan 430212, China
wst_itengineer@outlook.com

**Abstract.** Thermal infrared (TIR) images typically lack detailed features and have low contrast, making it challenging for conventional feature extraction models to capture discriminative target characteristics. As a result, trackers are often affected by interference from visually similar objects and are susceptible to tracking drift. To address these challenges, we propose a novel saliency-guided Siamese network tracker based on key fine-grained feature information. First, we introduce a fine-grained feature parallel learning convolutional block with a dual-stream architecture and convolutional kernels of varying sizes. This design captures essential global features from shallow layers, enhances feature diversity, and minimizes the loss of fine-grained information typically encountered in residual connections. In addition, we propose a multi-layer fine-grained feature fusion module that uses bilinear matrix multiplication to effectively integrate features across both deep and shallow layers. Next, we introduce a Siamese residual refinement block that corrects saliency map prediction errors using residual learning. Combined with deep supervision, this mechanism progressively refines predictions, applying supervision at each recursive step to ensure consistent improvements in accuracy. Finally, we present a saliency loss function to constrain the saliency predictions, directing the network to focus on highly discriminative fine-grained features. Extensive experiment results demonstrate that the proposed tracker achieves the highest precision and success rates on the PTB-TIR and LSOTB-TIR benchmarks. It also achieves a top accuracy of 0.78 on the VOT-TIR 2015 benchmark and 0.75 on the VOT-TIR 2017 benchmark.

[†]These authors contributed equally to this work.
[*]Corresponding author.

## 1     Introduction

Thermal infrared (TIR) target tracking is a significant research focus within the field of computer vision [1]. With ongoing advancements in uncooled TIR imaging technology, TIR sensors are becoming more compact and capable of capturing higher-resolution imagery, providing a solid foundation for the development of more effective TIR tracking algorithms. Compared to visible-light tracking, TIR tracking offers notable advantages, including all-weather operation and strong resilience to environmental interference. These characteristics make it particularly valuable for applications such as autonomous driving, video surveillance, and maritime rescue. However, despite these benefits, TIR tracking faces several inherent challenges. TIR targets typically exhibit low signal-to-noise ratios, poor contrast, limited texture details, and blurred edges. These factors significantly impair the performance of traditional feature extraction models, making it difficult to generate clear and discriminative target representations and thereby increasing the complexity of the tracking process.

To enhance the robustness of target tracking, early methods primarily relied on handcrafted features to represent targets, such as Histogram of Oriented Gradients (HOG), intensity histograms, Scale-Invariant Feature Transform (SIFT), and Local Binary Patterns (LBP). For example, Comaniciu et al. [2] utilized kernel histograms to model target feature distributions and applied a non-parametric density gradient ascent algorithm to iteratively locate the target. Paravati et al. [3] adopted genetic algorithms to evaluate the similarity between candidate regions and templates using cross-correlation on raw intensity features. Similarly, Cheng et al. [4] proposed a cascaded grayscale feature space, leveraging orthogonal filters to extract directional subspace features for target representation. However, TIR images typically lack strong edge and texture information, which limits the effectiveness of these manually designed features in capturing discriminative target representations. As a result, traditional methods often struggle to deliver reliable performance in TIR tracking. Motivated by the success of deep learning in visible-spectrum tracking, researchers have increasingly adopted convolutional neural networks (CNNs) for TIR tracking, owing to their powerful feature representation capabilities. For instance, Berg et al. [5] integrated deep features with a distribution field-based template matching approach, updating templates online and incorporating background context to enhance region selection. Liu et al. [6] employed a pre-trained VGG to extract multi-level deep features from TIR targets and combined these with correlation filters and convolutional features to construct multiple weak trackers, enhancing tracking accuracy through effective scale estimation.

Despite these advancements, deep learning models still encounter challenges when tracking targets that are visually or semantically similar. In addition, pre-trained models often struggle to extract highly discriminative features in such scenarios. To overcome these limitations, recent research has increasingly focused on fine-grained feature learning specifically tailored for TIR tracking. For example, MLSSNet [7] constructs a multi-level similarity model by integrating global semantic and local structural cues, utilizing an ensemble of sub-networks to adaptively fuse similarity information. MMNet [8] adopts a dual-layer architecture that jointly captures both discriminative and fine-grained contextual features through a non-local attention mechanism and a multi-task learning framework. Similarly, Yang et al. [9] proposed a fine-grained feature network enhanced with a diversity loss function and a mask suppression mechanism, designed to extract distinctive features from multiple local regions. These recent developments have significantly improved the appearance modeling of TIR targets, resulting in enhanced tracking accuracy and robustness under TIR conditions.

To address the challenge that traditional models struggle to extract discriminative features when the target closely resembles the background, we propose FGSGT, a novel saliency-guided Siamese network tracker that utilizes key fine-grained feature information. The main contributions of this paper are summarized as follows:

- We propose a key fine-grained information saliency-guided network to extract global contextual information from shallow layers, enhances spatial detail, and mitigates the loss of fine-grained information in residual connections.
- We introduce a multi-level feature fusion module that employs outer product operations and normalization techniques to integrate deep and shallow features.
- We propose a residual refinement block within the Siamese framework, applying residual learning and deep supervision to correct prediction errors in the saliency maps and iteratively refine predictions quality during training.

## 2    Methodology

### 2.1    Preview of FGSGT

The architecture of the proposed FGSGT is illustrated in Fig. 1(a), with ResNet-50 serving as the backbone network. However, the original design uses large stride values, as high as 32 pixels, which are not suitable for the dense prediction tasks required by Siamese networks. To address this, the strides in the Conv4 and Conv5 blocks are reduced to 8 pixels. Additionally, dilated convolutions are used to expand the receptive field, enabling the network to capture more comprehensive contextual information [10]. To maintain dimensional consistency, a $1 \times 1$ convolutional layer is added after each residual block to reduce the output channels to 256. The feature maps from Conv3, Conv4, and Conv5 are then processed by three independent FGSGT_RPN modules, each producing feature maps with uniform spatial resolution. These feature maps are aggregated using weighted summation followed by a weighted fusion layer, allowing for effective extraction and integration of multi-branch features across different layers.

The architecture of the FGSGT_RPN module is illustrated in Fig. 1(b), where the extracted multi-level features are denoted as $\mathcal{F}_3(z)$, $\mathcal{F}_4(z)$, and $\mathcal{F}_5(z)$. These features are then fused through an encoder module to create a unified representation. To improve spatial feature extraction, we introduce a Depth-wise Cross-Correlation Operation (DW-Corr), which computes correlations independently for each input channel. This allows the model to capture fine-grained spatial relationships within individual feature channels more effectively. In the final stage, the head module generates classification (CLS) scores and bounding box regression (Reg) values. This is achieved using a cross-correlation layer, followed by a series of fully convolutional layers. The final outputs are aggregated via weighted fusion to produce the final tracking results.

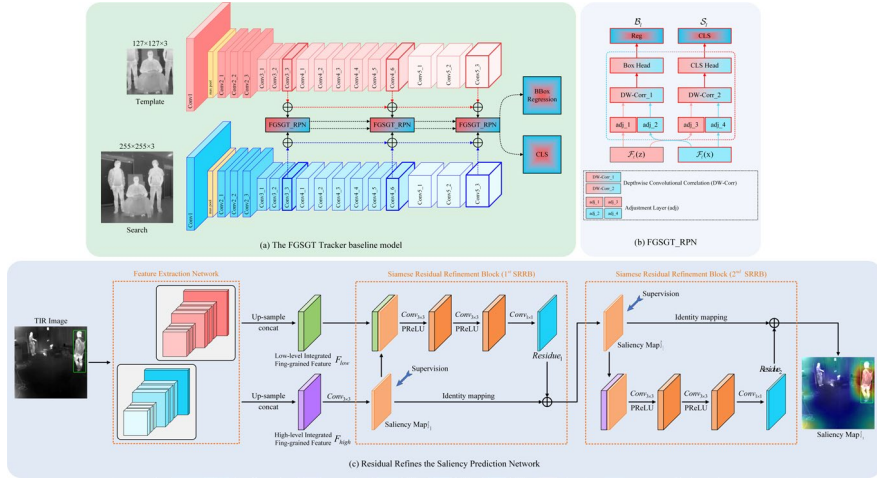

**Fig. 1.** The architecture of the proposed FGSGT tracker.

### 2.2   Key Fine-Grained Information Saliency-Guided Network

Thermal infrared (TIR) images typically have sparse texture details and low contrast, making it challenging for trackers to differentiate between the background and the target. Therefore, a supervised learning approach is necessary to effectively capture discriminative fine-grained TIR features. Given that saliency information can highlight target regions, we propose a key fine-grained information saliency-guided network, as shown in Fig. 1(c). First, fine-grained feature parallel learning convolutional blocks are employed to enhance feature diversity and prevent the loss of fine-grained information in residual connections. Second, a multi-layer fine-grained feature fusion model is proposed, where bilinear matrix multiplication is applied across different layers. This interlayer interaction aids in extracting discriminative features and improving feature representation. Third, a Siamese residual refinement block (SRRB) is introduced to correct prediction errors in the initial saliency map. Finally, a deep supervision mechanism is applied, where a series of SRRBs progressively refine the saliency predictions. Supervision signals are provided to the saliency maps at each recursive step during training, enabling iterative refinement of the saliency maps.

**Fine-grained feature parallel learning convolutional block.** The Conv4 layer of the ResNet-50 backbone network relies solely on the output from the final layer of the high-level feature extractor for classification, which overlooks crucial global feature information from the shallow layers. This limitation negatively impacts classification performance. To address this issue, we introduce the fine-grained feature parallel learning convolutional block (FGPCB). As shown in Fig. 2, FGPCB employs a dual-stream architecture with convolutional kernels of varying sizes to capture a diverse range of features. Downsampling is achieved by adjusting the kernel sizes and setting the stride to 2, which helps preserve fine-grained information in residual connections.

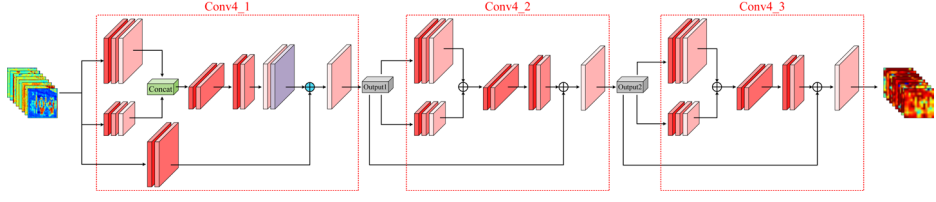

**Fig. 2.** Fine-grained feature parallel learning convolution block.

The parallel convolution operations in FGPCB play a critical role in extracting the region of interest. This structure concatenates different feature representation matrices using a Concat operation, increasing the channel dimension to enhance feature diversity. Additionally, FGPCB utilizes $1 \times 3$ and $3 \times 1$ convolutional kernels for feature extraction, while retaining the $Conv_{1 \times 1}$ convolution as a key element in the residual connection to ensure dimensional consistency. To reduce computational complexity, FGPCB replaces the $Conv_{3 \times 3}$ convolution with $1 \times 3$ and $3 \times 1$ convolutions. Furthermore, to align the dimensionality of feature information between deep and shallow structures, FGPCB removes the pooling layers and replaces the Concat operation with direct addition at the fourth layer.

To address the vanishing gradient problem, we replace the ReLU activation function with the Leaky ReLU activation function, defined as follows:

$$f(x) = \max(\alpha x, x), \quad (0 < \alpha < 1) \tag{1}$$

where $\alpha$ represents the gradient in the negative interval. In this paper, $\alpha$ is set to 0.25 to achieve optimal nonlinear fitting.

To enhance inter-layer information interaction, deep and shallow feature matrices are fused using matrix outer products and normalization operations. This fusion helps more accurately capture discriminative regions, thereby improving feature representation. Let $X \in \mathbb{R}^{H_x, W_x, C_x}$, $Y \in \mathbb{R}^{H_y, W_y, C_y}$, and $Z \in \mathbb{R}^{H_z, W_z, C_z}$ represent the multi-layer feature matrices extracted from the parallel convolutional block, defined as follows:

$$X = M(Conv_{3 \times 1}(Conv_{1 \times 3}(Concat(Conv_{1 \times 1}(F), Conv_{3 \times 3}(F))))) + Conv_{3 \times 3}(F) \tag{2}$$

$$Y = Conv_{3 \times 1}(Conv_{1 \times 3}(Conv_{1 \times 1}(X) + Conv_{3 \times 3}(X))) + X \tag{3}$$

$$Z = Conv_{3 \times 1}(Conv_{1 \times 3}(onv_{1 \times 1}(Y) + Conv_{3 \times 3}(Y))) + Y \tag{4}$$

where $Conv$ includes convolution, batch normalization, and activation layers; FF represents the feature map from the previous layer; $Conv_{1 \times 3}$, $Conv_{3 \times 1}$, $Conv_{1 \times 1}$, and $Conv_{3 \times 3}$ denote convolution kernels of sizes $1 \times 3$, $3 \times 1$, $1 \times 1$, and $3 \times 3$, respectively. Max pooling is used for downsampling to reduce the size of the feature map.

The model presented in our work is defined as:

$$O_{bp} = \sigma\left(N(\beta(X,Y)) + N(\beta(X,Z)) + N(\beta(Y,Z))\right) \tag{5}$$

where $\sigma$ represents the softmax function, $N$ denotes the normalization operation, and $\beta$ is the bilinear operator in each feature matrix.

To effectively fuse fine-grained features from different layers, we design a multi-layer fine-grained feature fusion model. This model improves feature map interactions across layers by using matrix outer products and normalization, projecting feature information from lower-dimensional to higher-dimensional space. The outputs of the last three layers are fused through matrix multiplication, then passed through a fully connected layer and the Softmax function, retaining the regions with the highest response values. The output feature matrix is defined as follows:

$$O = X^T \otimes Y + X^T \otimes Z + Y^T \otimes Z \tag{6}$$

Here, $\otimes$ denotes the bilinear operator between layers.

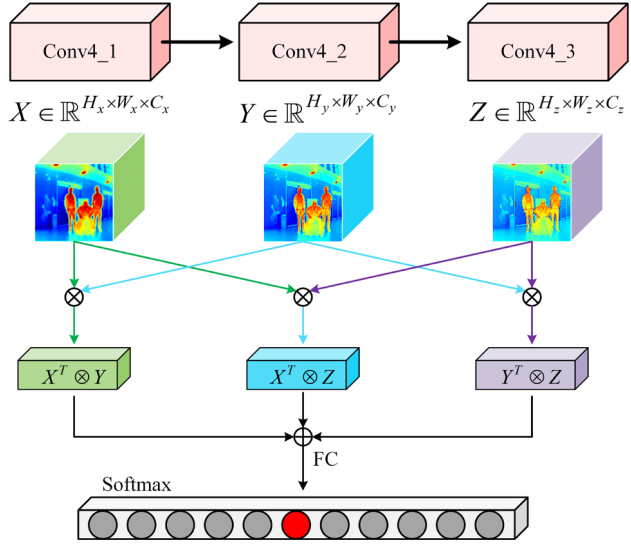

**Fig. 3.** Multi-layer fine-grained feature fusion model

The model architecture is shown in Fig. 3. The feature matrices $X, Y$, and $Z$ are obtained through the Conv4_1, Conv4_2, and Conv4_3 layers, respectively. The dimensions of all feature matrices are $(B, C, H \times W)$, where $B$ is the batch size, $C$ is the number of channels, and $H \times W$ is the spatial size of each feature map. Since inter-layer interactions help extract discriminative features, bilinear matrix multiplication is applied across different layers to enhance feature representation. Specifically, the matrix product of feature matrices $X$ and $Y$ is computed, followed by normalization to adjust the dimension of $X^T \otimes Y$ from $(B, C, C)$ to $(B, C \times C)$. This operation is repeated for $X$, $Y$, and $Z$, the three results are summed to form the input to the fully connected layer. In this way, fine-grained features from different layers are effectively fused.

**Siamese Residual Refinement Block.** We introduce a Siamese Residual Refinement Block (SRRB) to progressively reduce prediction errors in the saliency map, enabling step-by-step refinement. As shown in Fig. 1(c), the network architecture incorporates two SRRBs. Each SRRB alternately takes either the low-level integrated features $F_{low}$ or the high-level integrated features $F_{high}$, along with the saliency map from the previous iteration, as input. The refinement process calculates a residual, which is the difference between the ground truth and the saliency map from the previous step and then adds this residual to the previous saliency map to generate a more accurate saliency map. The refinement operation performed by the SRRB can be expressed as follows:

$$residue_i = \Phi_i(Cat(S_{i-1}), F) \quad (7)$$

$$S_i = S_{i-1} \oplus residue_i \quad (8)$$

In the $i$-th recursive step, the predicted saliency map $S_{i-1}$ from the previous step is concatenated with the feature map $F$. The function $\Phi$ is then applied to compute the residual $residue_i$. Finally, the residual $residue_i$ is added element-wise to $S_{i-1}$, and the updated saliency map $S_{i-1}$ is computed.

**Residual Refinement Saliency Prediction Network.** To effectively learn salient regions during the refinement process, we progressively enhance saliency predictions using a series of SRRBs. A set of multi-scale feature maps is generated by the fine-grained feature extraction module. The deep feature maps capture high-level semantic information of the salient target, while the shallow feature maps preserve fine structural details of the salient regions. The output feature maps from the convolutional layers are grouped into two categories: low-level feature maps, comprising Conv1, Conv2, and Conv3; and high-level feature maps, comprising Conv4 and Conv5.

The feature maps from Conv2 and Conv3 are upsampled to match the size of the Conv1 feature map. Then, through concatenation and convolution operations, redundant channels are reduced to generate the low-level integrated feature $F_{low}$:

$$F_{low} = f_{conv}(Cat(F_1, F_2, F_3)) \quad (9)$$

Here, $F_i$ represents the upsampled feature map from the i-th convolutional layer; $Cat(\cdot)$ denotes concatenation of the feature maps from the first three layers; and $f_{conv}$ refers to the feature fusion network, consisting of three convolutional layers and a PReLU activation function.

Similarly, the high-level integrated feature $F_{high}$ is defined as:

$$F_{high} = f_{conv}(Cat(F_4, F_5)) \quad (10)$$

First, the network predicts the saliency map $S_0$ from the high-level integrated feature $F_{high}$. This map captures the location of the salient target but lacks fine-grained saliency details. Next, starting from $S_0$, a series of SRRBs are applied to progressively refine the saliency prediction. Since the low-level integrated feature $F_{low}$ provides finer saliency details, the first SRRB is constructed by setting the feature map $F$ to $F_{low}$ (as shown in equation (7), refining $S_0$ to produce a more detailed saliency map $S_1$. However, $F_{low}$ also includes substantial non-salient information, which can introduce irrelevant regions into $S_1$. To address this, the second SRRB is constructed by replacing $F_{low}$ with $F_{high}$ in equation (7), which helps eliminate the non-salient regions introduced by $F_{low}$. Since $F_{high}$ captures the semantic features of the salient target, this operation effectively removes non-salient details outside the semantic regions.

The algorithm steps are as follows:

---

**Algorithm 1:** Saliency Refinement Using SRRB
---
**Input:** Initial saliency map $S_0$; Low-level feature map $F_{low}$;
   High-level feature map $F_{high}$;
Initialize $S_0$ from high-level feature map $F_{high}$
**for** each iteration $i = 1$ to $N$ **do**
  **if** $i$ is odd **then**
    set $F_{input} \leftarrow F_{low}$;
  **else**
    set $F_{input} \leftarrow F_{high}$
**end if**
  Compute $residue_i$ according to Eq.(7)
  Update saliency map $S_i$ according to Eq.(8)
  Apply supervision to $S_i$ during training
**end for**
**Output:** Final refined saliency map $S_{final} \leftarrow S_N$

---

To further improve the accuracy of saliency prediction, we utilize a series SRRBs that iteratively refine the prediction by alternately combining high-level features $F_{high}$ and low-level features $F_{low}$. A deep supervision mechanism is also employed, where supervision signals are applied to the saliency map at each recursive step during training. As shown in Fig. 1(c), the inclusion of auxiliary supervision at intermediate stages enables each SRRB to directly learn residuals relative to the ground truth, simplifying the network optimization process. The final saliency map, $S_{final}$, produced at the last recursive step, is taken as the final output of the network.

**FGSGT Loss Function.** The proposed loss function consists of three components: classification loss, regression loss, and saliency prediction loss.

The total loss function is expressed as:

$$\mathcal{L}_{FGSGT} = \lambda_{cls}\mathcal{L}_{cls} + \lambda_{reg}\mathcal{L}_{reg} + \lambda_{sal}\mathcal{L}_{sal} \tag{11}$$

where $\lambda_{cls}$, $\lambda_{reg}$ and $\lambda_{sal}$ represent the weights of each component loss.

The network generates multiple saliency maps, including the initial saliency map $S_0$, as well as a sequence of refined saliency maps $(S_1, ..., S_N)$ through $N$ SRRBs. During training, a deep supervision mechanism is employed, where supervision signals are applied to each saliency map, and the cross-entropy loss between each predicted saliency map and the ground truth is computed. Therefore, the saliency prediction loss $\mathcal{L}_{sal}$ is defined as the sum of the losses for all predicted saliency maps:

$$\mathcal{L}_{sal} = w_0 y_0 + \sum_{i=1}^{N} w_i y_i \tag{12}$$

Here, $w_0$ and $y_0$ represent the weight and loss of the initial saliency prediction, respectively, while $w_i$ and $y_i$ denote the weights and losses of the $i$-th recursive step. $N$ is the number of recursive steps for SRRB refinement.

The classification loss $\mathcal{L}_{cls}$ is defined using the cross-entropy loss as follows:

$$\mathcal{L}_{cls} = -\frac{1}{2}\sum_j \left[u_j^* \times \log u_j + (1 - u_j^*) \times \log(1 - u_j)\right] \tag{13}$$

where $u_j$ represents the probability predicted by the network, and $u_j^*$ is the true value for the $j$-th sample. The regression loss $\mathcal{L}_{reg}$ uses the L1 loss and is defined as follows:

$$L_{reg} = \frac{1}{N_{pos}}\sum_j \left[u_j^* > 0\right]\left\|\lambda_G \times \mathcal{L}_{IoU}(p_j, \hat{p}_j) + \lambda_1 \times \mathcal{L}_1(p_j, \hat{p}_j)\right\|_1 \tag{14}$$

where $N_{pos}$ is the number of positive samples, $\left[u_j^* > 0\right]$ is an indicator function that equals 1 when $u_j^* > 0$ and 0 otherwise, $\mathcal{L}_{IoU}$ is the IoU loss, $\mathcal{L}_1(\cdot)$ is the L1 norm loss. $\lambda_G$ and $\lambda_1$ are the regularization parameters. $p_j$ represents the predicted bounding box for the $j$ sample; and $\hat{p}_j$ denotes the ground truth bounding box label.

## 3    Experiment

### 3.1    Implementation Details

**Dataset and Evaluation Criteria.** The PTB-TIR benchmark [11] is a TIR pedestrian dataset designed for short-term tracking tasks, consisting of 60 manually annotated TIR pedestrian sequences with a total of over 30K frames. The LSOTB-TIR benchmark [12] is currently the largest TIR dataset, encompassing 47 object types and 1,400 TIR video sequences with over 600K frames. It also provides annotations for 730K bounding boxes. Following the methods outlined in [11] and [12], we adopt center location error (Precision) and overlap score (Success) as evaluation metrics for the PTB-TIR benchmark. Additionally, Normalized Precision (NP) is introduced as an evaluation metric for the LSOTB-TIR benchmark. The VOT-TIR2015 [13] and VOT-TIR2017 [14] benchmarks are short-term TIR object tracking datasets published by the Visual Object Tracking (VOT) committee. These benchmarks contain 45 test sequences, spanning eight object categories, with an average sequence length of 651 frames. As described in [13] and [14], we use Accuracy, Robustness, and Expected Average Overlap (EAO) as the evaluation metrics for these two benchmarks.

**Experimental Platform.** The proposed FGSGT tracker is implemented using Python 3.9 and trained on the LSOTB-TIR TIR dataset, along with the visible light datasets VOT 2017 and VOT 2019. The training process is accelerated using CUDA 11.8 and cuDNN 7.6. The search region has an image block size of 255×255, while the template region has an image block size of 127×127. All experiments are conducted on a 24GB NVIDIA GeForce RTX 3090 GPU.

### 3.2    Ablation Experiment

We conduct ablation experiments on PTB-TIR and LSOTB-TIR benchmarks to assess the effectiveness of the proposed Fine-Grained Feature Parallel Learning Convolutional Block (FGPCB), the Residual Refinement Saliency Prediction Network (RRSPN), and the FGSGT loss function. The results of these ablation experiments are presented in Table 1, where SiamRPN++ serves as the baseline tracker.

**Table 1.** Ablation results on PTB-TIR and LSOTB-TIR benchmarks

| Model | Component | | | PTB-TIR | | LSOTB-TIR | | |
|---|---|---|---|---|---|---|---|---|
| | FGPCB | RRSPN | Loss | Pre.↑ | Suc.↑ | Pre.↑ | Norm. Pre.↑ | Suc.↑ |
| SiamRPN++(base) | | | | 0.740 | 0.565 | 0.742 | 0.695 | 0.584 |
| Siam_FGPCB | √ | | | 0.775 | 0.592 | 0.769 | 0.712 | 0.615 |
| Siam_RRSPN | | √ | | 0.770 | 0.605 | 0.762 | 0.709 | 0.628 |
| Siam_T | √ | √ | | 0.795 | 0.615 | 0.825 | 0.734 | 0.639 |
| FGSGT (ours) | √ | √ | √ | **0.830** | **0.625** | **0.845** | **0.762** | **0.652** |

As shown in Table 1, the Siam_FGPCB model outperforms the baseline tracker SiamRPN++ in overall performance. On the PTB-TIR dataset, it achieves improvements of 3.5% in precision and 2.7% in success rate. On the LSOTB-TIR dataset, it demonstrates gains of 2.7% in precision, 1.7% in normalized precision, and 3.1% in success rate. These results highlight the effectiveness of integrating the fine-grained feature parallel learning convolutional block to enhance tracking performance. In addition, the Siam_RRSPN model shows improvements of 3.0% and 2.0% in precision, and 4.0% and 4.4% in success rate on the PTB-TIR and LSOTB-TIR datasets, respectively, validating the effectiveness of the residual refinement saliency prediction network. Furthermore, the Siam_T model, built upon Siam_RRSPN, achieves further performance improvements, with precision and success rate gains of 2.5% and 1.0%, respectively, on the PTB-TIR dataset. On the LSOTB-TIR dataset, it achieves improvements of 6.3% in precision, 2.5% in normalized precision, and 1.1% in success rate.

### 3.3    Performance Comparison with State-of-the-Arts

To evaluate the proposed FGSGT, we compared it with various state-of-the-art trackers, and experimental results on LSOTB-TIR, PTB-TIR, VOT-TIR2015, and VOT-TIR2017 benchmarks are presented in Fig. 4, Fig. 5, and Table 2, respectively.

**Results on PTB-TIR and LSOTB-TIR.** As shown in Fig. 4, the proposed FGSGT tracker achieves the best performance across all evaluation metrics on the PTB-TIR benchmark, with a precision score of 0.830 and a success rate of 0.625. Compared to the baseline tracker SiamRPN++, FGSGT improves precision and success rate by 9.0% and 6.0%, respectively. Additionally, when compared to Transformer-based trackers such as TransT and DFG, FGSGT demonstrates improvements of 2.1% and 1.5% in precision, respectively. In comparison with correlation filter-based trackers like ECO-stir and MCCT, FGSGT shows gains of 2.6% and 8.5% in success rate, respectively.

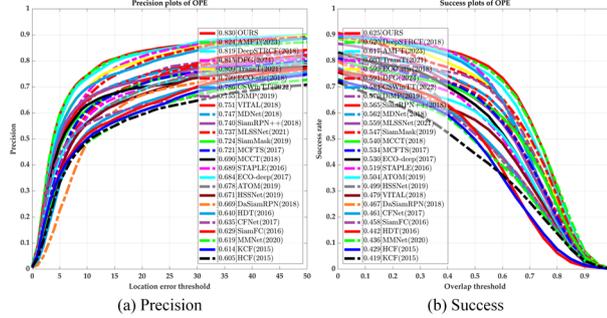

(a) Precision   (b) Success

**Fig. 4.** Performance comparison results on PTB-TIR benchmark.

**Results on LSOTB-TIR.** As shown in Fig. 5, FGSGT achieves a precision of 0.845, surpassing AMFT, DFG, and TransT. Specifically, FGSGT outperforms DFG by 1.1% in precision and 0.9% in success rate. Moreover, FGSGT reaches the highest success rate of 0.652. Compared to deep learning-based trackers such as DiMP and VITAL, the proposed tracker improves the success rate by 5.0% and 21.4%, respectively.

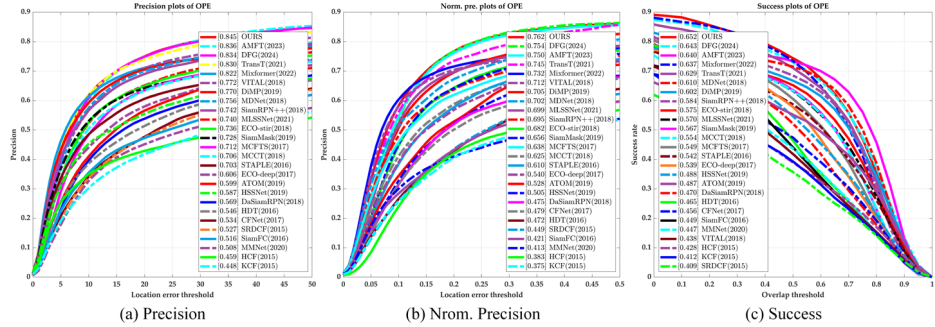

(a) Precision   (b) Nrom. Precision   (c) Success

**Fig. 5.** Performance comparison results on LSOTB-TIR benchmark.

**Results on VOT-TIR 2015 and VOT-TIR 2017.** As shown in Table 2, the proposed FGSGT tracker achieves a precision score of 0.78 on the VOT-TIR2015 benchmark, tying for first place with DFG. On the VOT-TIR2017 benchmark, it achieves the highest precision score of 0.75, surpassing methods such as SiamSAV, Ocean, and TransT. Additionally, FGSGT obtains an EAO score of 0.375 on the 2015 benchmark, exceeding SiamRPN++ and ranking second only to UDCT. Compared to the baseline tracker SiamRPN++, FGSGT improves precision by 4.0% and 6.0% on the 2015 and 2017 benchmarks, respectively, and increases the EAO by 6.2% and 6.4%, respectively. Finally, the robustness of FGSGT reach 1.59 and 1.90 on the 2015 and 2017 benchmarks, respectively, outperforming Siamese network-based trackers such as MLSSNet and MMNet, as well as the Transformer-based tracker DFG.

Table 2. Tracking challenges results on LSOTB-TIR benchmark

| Methods | Trackers | VOT-TIR 2015 | | | VOT-TIR 2017 | | |
|---|---|---|---|---|---|---|---|
| | | EAO↑ | Acc↑ | Rob↓ | EAO↑ | Acc↑ | Rob↓ |
| Correlation filter | SRDCF | 0.225 | 0.62 | 3.06 | 0.197 | 0.59 | 3.84 |
| | HDT | 0.188 | 0.53 | 5.22 | 0.196 | 0.51 | 4.93 |
| | ECO-deep | 0.286 | 0.64 | 2.36 | 0.267 | 0.61 | 2.73 |
| | MCFTS | 0.218 | 0.59 | 4.12 | 0.193 | 0.55 | 4.72 |
| | MCCT | 0.250 | 0.67 | 3.34 | 0.270 | 0.53 | 1.76 |
| | ECO-MM | 0.303 | 0.72 | 2.44 | 0.291 | 0.65 | 2.31 |
| | ECO_LS | 0.319 | 0.64 | 0.82 | 0.302 | 0.55 | 0.93 |
| | ECOHG_LS | 0.270 | 0.60 | 0.92 | 0.251 | 0.49 | 1.26 |
| Transformer | TransT | 0.287 | 0.77 | 2.75 | 0.290 | 0.71 | 0.69 |
| | DFG | 0.329 | **0.78** | 2.41 | 0.304 | 0.74 | 2.63 |
| | CorrFormer | 0.269 | 0.71 | 0.56 | 0.262 | 0.66 | 1.23 |
| Deep learning | CREST | 0.258 | 0.62 | 3.11 | 0.252 | 0.59 | 3.26 |
| | DeepSTRCF | 0.257 | 0.63 | 2.93 | 0.262 | 0.62 | 3.32 |
| | VITAL | 0.289 | 0.63 | 2.18 | 0.272 | 0.64 | 2.68 |
| | DiMP | 0.330 | 0.69 | 2.23 | 0.328 | 0.66 | 2.38 |
| | ATOM | 0.331 | 0.65 | 2.24 | 0.290 | 0.61 | 2.43 |
| | Ocean | 0.339 | 0.70 | 2.43 | 0.320 | 0.68 | 2.83 |
| Siamese network | SiamFC | 0.219 | 0.60 | 4.10 | 0.188 | 0.50 | **0.59** |
| | CFNet | 0.282 | 0.55 | 2.82 | 0.254 | 0.52 | 3.45 |
| | DaSiamRPN | 0.311 | 0.67 | 2.33 | 0.258 | 0.62 | 2.90 |
| | SiamRPN | 0.267 | 0.63 | 2.53 | 0.242 | 0.60 | 3.19 |
| | TADT | 0.234 | 0.61 | 3.33 | 0.262 | 0.60 | 3.18 |
| | HSSNet | 0.311 | 0.67 | 2.53 | 0.262 | 0.58 | 3.33 |
| | SiamRPN++ | 0.313 | 0.74 | 2.25 | 0.296 | 0.69 | 2.63 |
| | MMNet | 0.340 | 0.61 | 2.09 | 0.320 | 0.58 | 2.90 |
| | MLSSNet | 0.329 | 0.57 | 2.42 | 0.286 | 0.56 | 3.11 |
| | SiamSAV | 0.344 | 0.71 | 2.75 | 0.336 | 0.67 | 2.92 |
| | UDCT | **0.420** | 0.67 | 0.88 | 0.342 | 0.66 | 0.81 |
| | FGSGT (ours) | 0.375 | **0.78** | 1.59 | **0.360** | **0.75** | 1.90 |

### 3.4   Visualization of Tracking Results

To provide a more intuitive demonstration of the effectiveness of the proposed FGSGT tracker, Fig. 6 and Fig. 7 illustrate the visual tracking results of FGSGT compared to AMFT, TransT, and DeepSTRCF on six challenging sequences from the LSOTB-TIR and PTB-TIR benchmarks. In these figures, the green line represents the ground truth bounding box of the target.

As shown in Fig. 6 and Fig. 7, FGSGT achieves more accurate tracking results in most of the sequences. Specifically, in the "birds" and "classroom2" sequences shown in Fig.6(a) and Fig. 6(b), the presence of distractor objects causes AMFT and TransT to exhibit tracking failure or drift at frames 65 and 224, respectively. In contrast, FGSGT, by learning highly discriminative fine-grained features of the target, distinguishes between the distractor and the actual target, enabling accurate localization.

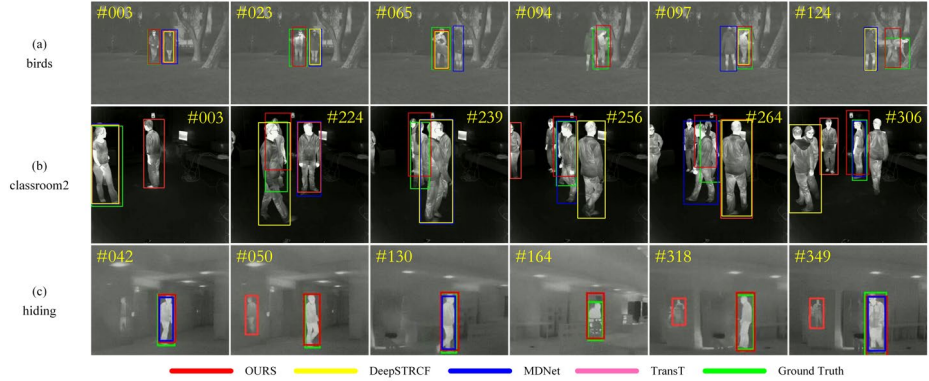

**Fig. 6.** Visualization of tracking results on PTB-TIR benchmark.

Similarly, in the "person_D_004" sequence shown in Fig. 7(a), background interference from the water surface and partial occlusion lead to tracking drift for AMFT and TransT at frame 606, and target loss for DeepSTRCF at frame 657. However, FGSGT is still able to accurately locate the target. Furthermore, Fig. 7(b) and Fig. 7(c) present the "person_H_002" and "person_S_018" sequences, which are characterized by low resolution and poor signal-to-noise ratio, resulting in weak target feature representation and the presence of multiple similar objects. These challenges cause DeepSTRCF, TransT, and AMFT to lose the target at frames 39 and 41, respectively. Due to its superior target representation capability, FGSGT demonstrates more accurate localization in these sequences.

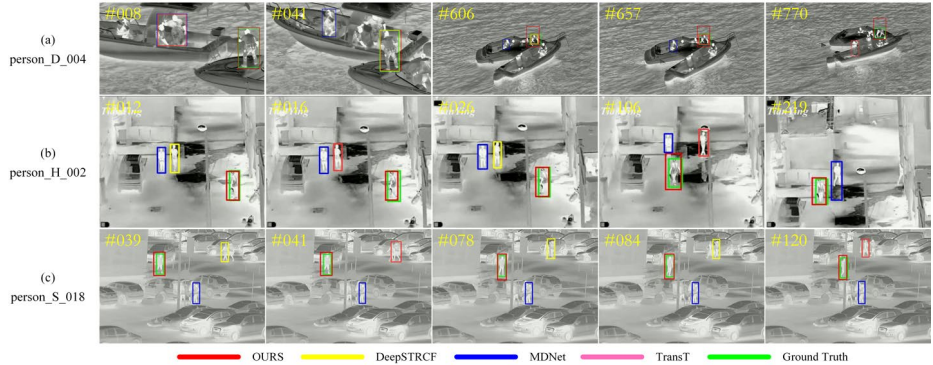

**Fig. 7.** Visualization of tracking results on LSOTB-TIR benchmark.

## 4    Conclusions

In this paper, we propose FGSGT, a saliency-guided Siamese network tracker based on key fine-grained feature information, to address the challenges of extracting discriminative features from TIR targets. These challenges are often due to low contrast and sparse feature details in TIR images, which lead traditional trackers to suffer from interference by visually similar objects and tracking drift. Our tracker introduces several key innovations, including a fine-grained feature parallel learning convolutional block with a dual-stream architecture and varying convolutional kernel sizes, which enhances feature diversity and preserves fine-grained details. We also propose a multi-layer feature fusion module using bilinear matrix multiplication to effectively integrate deep and shallow features. The Siamese residual refinement block refines saliency map predictions through residual learning and deep supervision, improving accuracy. Finally, a dedicated saliency loss function guides the network to focus on highly discriminative fine-grained features. Extensive experimental results on the PTB-TIR, LSOTB-TIR, and VOT-TIR benchmarks demonstrate that FGSGT outperforms existing trackers, achieving higher precision, success rates, and robustness.